# A platform for causal knowledge representation and inference in industrial fault diagnosis based on cubic DUCG

**Bu Xusong[1], Nie Hao[2](co-first author), Zhang Zhan[2], Zhang Qin[1,2,*]**

[1] Department of Computer Science and Technology, Tsinghua University, Beijing 100084, China; bux-usong@mail.tsinghua.edu.cn; qinzhang@mail.tsinghua.edu.cn
[2] Institute of Nuclear and New Energy Technology, Tsinghua University, Beijing 100084, China; nh17@mails.tsinghua.edu.cn; zhangzhan19@tsinghua.edu.cn
[*] Correspondence: qinzhang@mail.tsinghua.edu.cn

**Abstract:** The working conditions of large-scale industrial systems are very complex. Once a failure occurs, it will affect industrial production, cause property damage, and even endanger the workers' lives. Therefore, it is important to control the operation of the system to accurately grasp the operation status of the system and find out the failure in time. The occurrence of system failure is a gradual process, and the occurrence of the current system failure may depend on the previous state of the system, which is sequential. The fault diagnosis technology based on time series can monitor the operating status of the system in real-time, detect the abnormal operation of the system within the allowable time intervals, diagnose the root cause of the fault and predict the status trend. In order to guide the technical personnel to troubleshoot and solve related faults, in this paper, an industrial fault diagnosis system is implemented based on the cubic DUCG theory. The diagnostic model of the system is constructed based on expert knowledge and experience. At the same time, it can perform real-time fault diagnosis based on time sequence, which solves the problem of fault diagnosis of industrial systems without sample data.

## 1. Introduction

Industrial fault diagnosis is the technology for early fault detection, as well as for forecasting the development trend of faults by monitoring real-time abnormal signals[1-3]. The fault behavior of large complex systems is usually dynamic, and the signals received from online measuring points have the characteristics of time sequence and fluctuation[4-11]. Therefore, a time-series modeling algorithm is needed to solve the real-time estimation and prediction of industrial systems. General algorithms employed for time-series fault diagnosis are based on statistical learning models, including the Hidden Markov model (HMM)[12-14], Kalman filter model (KFM)[15-17], and Dynamic Bayesian Network model(DBNs)[18,19]. Due to the high complexity of the industrial system and higher data dimension, when these algorithms are applied to the fault diagnosis of complex industrial systems, there are some problems, such as a large amount of calculation, long response time, and lack of ability to deal with incomplete and uncertain information. HMM is memoryless and cannot utilize context information, because it is only related to its previous state. If you want to use more known information, you must establish a high-order HMM model. Kalman filter can accurately estimate the linear process model and measurement model, but can not achieve the optimal estimation effect in the nonlinear scene. In order to set the linear environment, it is necessary to assume that the process model is a constant speed model, however, in practical application, both the process model and the measurement model are nonlinear. The precise reasoning calculation of DBNs is an NP-hard problem[20,21], this affects its reasoning efficiency. Meanwhile, it is difficult to construct complex Bayesian Networks, those problems hinder the practical application of DBNs in complex industrial fault diagnoses. In addition, these models are statistically based, the training of these models requires a large number of high-quality data sets, but for some industrial systems without a lot of high-quality data, these methods can not be applied. Such as the nuclear power plant system, and spacecraft in orbit, few fault data can be used for model training[22]. Therefore, a time-series fault diagnosis system with little or no data dependence, explainable, computationally efficient, and easy to implement is necessary.

Dynamic Uncertain Causality Graph (DUCG) is a probability graph model[23]. It can intuitively describe the uncertain causality of events by probabilities and different graphic symbols and has a strong ability to express the propagation of causal uncertainty, the advantages of visualization, interpretability, and high computational efficiency. The DUCG has a complete mathematical foundation and theoretical system and provides a concise expression and reasoning method of uncertain knowledge in the form of causal graphs. It can realize the causal inference problems of discrete, continuous, and fuzzy variables. At the same time, it can handle causal loops, allowing directed loops in the model[24-27]. These features enable DUCG to efficiently implement accurate inference of multi-connected causality. The DUCG's reasoning knowledge base can be constructed by domain experts based on their knowledge and experience[28-30]. Therefore, it is suitable for scenarios where causal inference is required, but there is no data to build a model.

In this paper, we build an industrial fault diagnosis platform based on cubic DUCG theory. It supports experts to model diagnostic faults based on experience and knowledge, and based on the model, implements time series-based fault diagnosis and prediction. The following sections of this paper are arranged as follows: Section 2 is the introduction of DUCG theory and the description of the reasoning algorithm used in this diagnosis system. Section 3 is the design and implementation of the diagnosis system. Section 4 is system verification. Section 5 summarizes the paper and suggests directions for future research.

## 2. DUCG Theory and the Inference Algorithm of cubic DUCG

The DUCG is a hybrid model, it has two sub-models. One is the single-valued DUCG (S-DUCG) completely obeying the probability theory[31]. Another is the multi-valued DUCG (M-DUCG) somehow exceeding probability theory. The so-called single-valued means that only the causes of the true state of a variable can be represented, while the false state must be the complement of the true state. M-DUCG does not have this limitation[23].

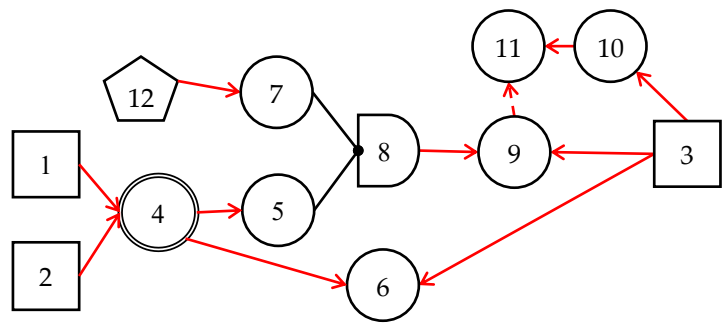

**Figure 1**. An example of the M-DUCG model. It describes the fault propagation process by graphic symbols. The numbered squares 1-3 stands for three root causes (faults), other shapes stand for some abnormal signals that may be caused by them.

The M-DUCG describes the conditional independent relation between multi-valued random variables by graph structures, which brings great convenience to the research of the probability model in a high dimensional space. The M-DUCG model can be described as $G(V, R)$, $V=(B_i, X_k, BX_j, G_n, D_m)$ is the collection of variables. $R=(F_{ij},$ conditional $F_{ij})$ is the collection of relations between different variables. For example, $F_{ij}$ indicates the causal relation from $V_i$ to $V_j$, and causality propagates from parent to child. In M-DUCG, different types of variables and relations are represented by different shapes. Figure 1 shows a model of M-DUCG. It describes the causal relations of variables, and different types of variables are indicated by different shapes. For example, $B_1$ is a *B*-type variable numbered 1, it is represented as . The physical meanings of variables in the M-DUCG are shown in Table 1. Different shapes are used to construct the DUCG knowledge base, which can make the knowledge base clearer, easier to understand and facilitate the subsequent maintenance of the knowledge base.

**Table 1.** The physical meanings of variables in M-DUCG.

| Type | Shape | Description |
|---|---|---|
| $B_i$ |  | The root cause variable or the hypothesis. It has no parent but at least one child. |
| $X_i$ | 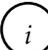 | The consequence variable, can also be used as the cause of other variables. |
| $BX_i$ |  | The integrated cause variable. It is used to represent the integrated effect of a group of *B*-type variables. |
| $G_i$ |  | The logic gate variable. It is used to describe the logical relation combination of parent variables. |
| $D_i$ |  | The default cause variable, and the default or unspecified cause of $X_i$ is defined as variable "$D_i$". |
| $F_{ij}$ | 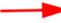 | $F_{i;j} = \left( r_i / r_n \right) A_{i;j}$ is the weighted functional event, indicates the causal relation from the parent variable $V_i$ to the child variable $V_j$. ($r_i / r_n$) is the weight. $A_{i;j}$ is the functional event, it records the intensity of causal effect between parent and child variables. |
| conditional $F_{ij}$ | 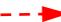 | The conditional causal relation from $V_i$ to $V_j$. Its condition is described as $Z_{ij}$. When $Z_{ij}$ is true, the causal relation holds. Otherwise, there is no causal relation between $V_i$ and $V_j$. |

The inference process of M-DUCG contains four steps: DUCG simplification, evidence expansion, getting hypothesis space, and probability calculation[23]. Similar to other probability models, the inference of the DUCG can be transformed into solving posterior probability problems. The conditional probability distribution of the DUCG is:

$$\Pr\left\{ H_{nk} \mid E \right\} = \frac{\Pr\left\{ H_{nk} E \right\}}{\sum\limits_{n,k} \Pr\left\{ H_{nk} E \right\}} \tag{1}$$

In equation (1), $\Pr\left\{ H_{nk} E \right\}$ is the probability of the *sub-DUCG_n* which only includes the hypothesis $H_{nk}$. $H_{nk}$ is the hypothesis, n is its index, indicates different hypotheses, and $k$ is its state, usually, the *B*-type or *BX*-type variable is used as the hypothesis. $\sum_{n,k} \Pr\left\{ H_{nk} E \right\}$ is the sum of probabilities of all sub-DUCGs. $E = E'E'' = \prod X_{nk}$ is the observed evidence, $E'$ is the abnormal evidence and $E''$ is the normal evidence. Before probability calculations, the logical expansion of the observed evidence should be carried out, and the essence of logical expansion is to recursively express child variables with parent variables. All the observed evidence should be expanded along its causal chain to the root cause events related to them. The evidence expansion expression is shown in Equation (2), and Figure 2 is the explanation for Equation (2).

$$X_{nk} = \sum_i F_{nk;i} V_i = \sum_i \left( r_{n;i} / r_n \right) \sum_j A_{nk;ij} V_{ij} \tag{2}$$

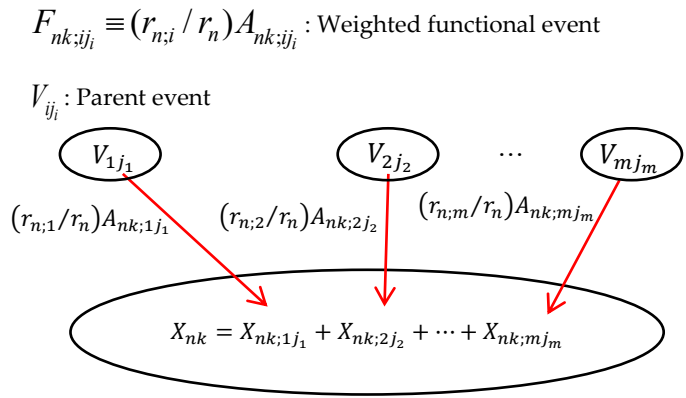

**Figure 2.** The illustration to the expression expansion in an M-DUCG model. The child event is expressed by its parent events.

As shown in Equation (2) and Figure 2, any $(r_{n;i}/r_n)A_{nk;ij}$ causes $X_{nk;ij}$ individually, resulting in $X_{nk}$ equally, where $A_{nk;ij}$ are in the newly defined *OR* relationship that is similar to exclusive *OR*, but is not the same completely, because of the weight $(r_{n;i}/r_n)$. We call this model as weighted set theory and the operator "+" in M-DUCG as weighted exclusive *OR*[32]. By using the evidence expansion, the child events can recursively be expanded to their parent events, until recursively expanded to the root cause event. After the logical expansion, the observed evidence will be in the form of sum-of-products composed of only {$B$-, $A$-, $r$-}-type events and parameters, and the resulting expression is used to calculate the conditional probability by the Equation (1).

The cubic DUCG is a theoretical model developed from M-DUCG proposed for the complex situation of dynamic negative feedback situation. The cubic DUCG inherits the graphical variable expression in M-DUCG, and the construction method of the cubic DUCG knowledge base is the same as that of M-DUCG. The main difference is that the time dimension is added to generate cubic DUCG, to more accurately express the logical relationship of events with time[33,34]. The inference method of cubic DUCG is shown as follows:

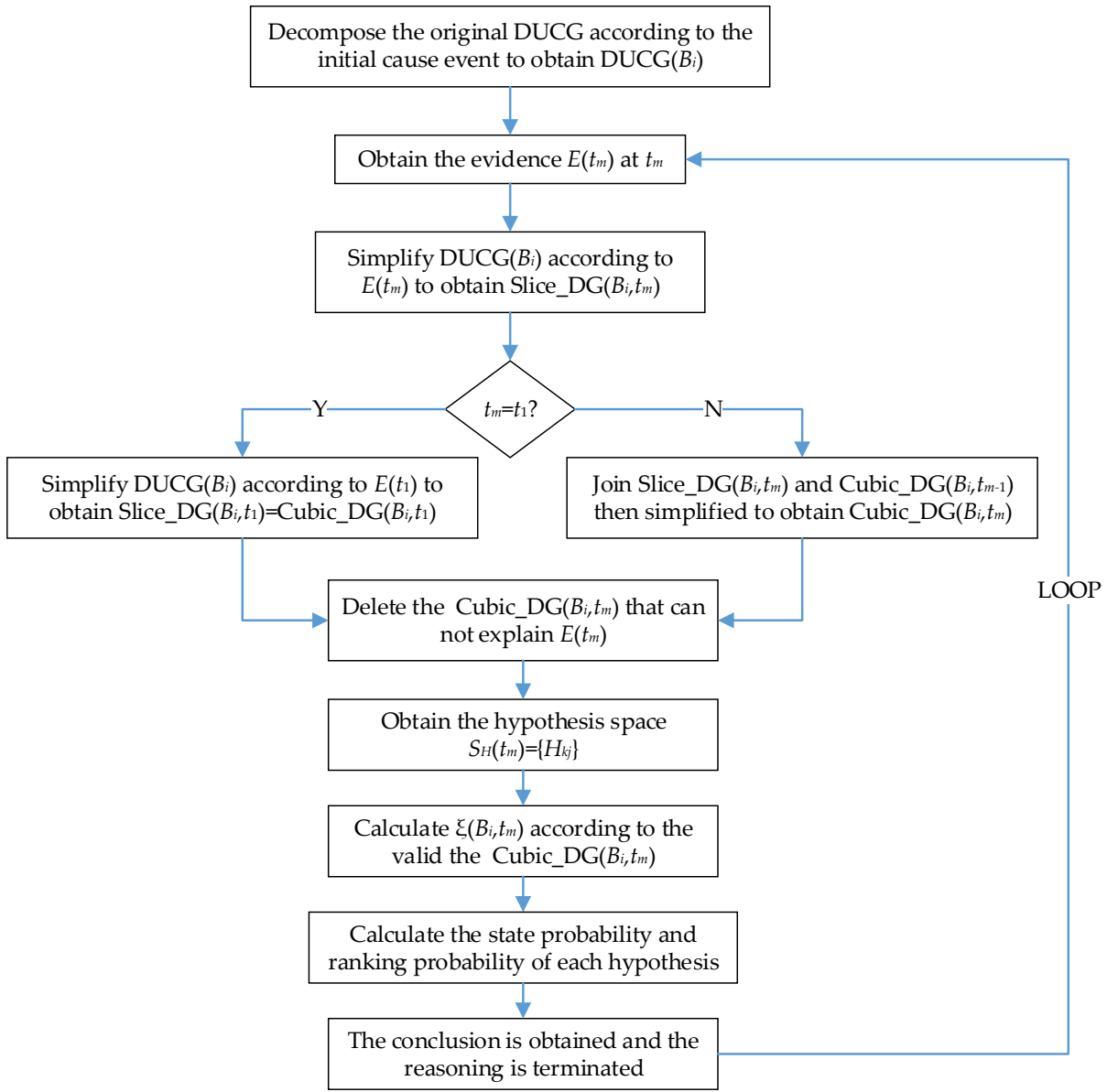

**Figure 3.** The flow chart of modeling and reasoning about the cubic DUCG. *Slice_DG(t_m)* is an intraslice causality graph at time $t_m$. *Cubic_DG(B_i, t_m)* is a cubic causality graph related to a specific fault mode (indexed by $B_i$) at $t_m$.

The flow chart of modeling and reasoning method based on cubic DUCG is shown in Figure 3, and according to this flow chart, the reasoning step of the cubic DUCG is shown as follows:

1. DUCG decomposition. The original DUCG is decomposed into several sub-DUCGs, each sub-DUCG only contains one initial event $B_i$, remarked as *DUCG(B_i)*. The purpose of DUCG decomposition is to reduce reasoning complexity and improve reasoning efficiency.

2. *DUCG(B_i)* simplification. According to the evidence $E(t_m)$ at $t_m$, *DUCG(B_i)* is simplified by the cubic DUCG simplification rules. The simplified *DUCG(B_i)* is called *Slice_DG(B_i, t_m)*, it describes the causality between initial event $B_i$ and the evidence $E(t_m)$ at $t_m$.

3. *Cubic_DG(Bi,tm)* generation. The *Cubic_DG(tm)* is generated by merging the *Cubic_DG(Bi,tm-1)* generated at *tm-1* and *Slice_DG(Bi,tm)* obtained at *tm*. In particular, when *tm*=1, *Cubic_DG(Bi,tm)= Slice_DG(Bi,tm)*. During the merging process, the causal linkage operation is done to connect the same variable between different time slices. For the generated *Cubic_DG(Bi,tm)*. The *Cubic_DG(Bi,tm)* is used for probability calculation, if it can not explain all the evidence *E(tm)* at *tm*, it is regarded as an invalid *Cubic_DG(Bi,tm)* and discarded. Because only the *Cubic_DG(Bi,tm)* can explain all abnormal evidence *E(tm)* is regarded as a valid *Cubic_DG*, the *Bi* in valid *Cubic_DG(Bi,tm)* is regarded as the hypothesis *Hkj*, which consists of the hypothesis space *SH(tm)*, *SH(tm)*={*Hkj*}.

4. If there is more than one hypothesis in *SH(tm)*, the state probability and ranking probability of each *Hkj* is calculated to evaluate which *Hkj* are more likely to happen. The probability is calculated by Equation (3), *E(tm)* is the abnormal evidence at time *tm*,

$E\left(t_m\right)=\prod_i X_{i,j_i}$ . $\Pr\left\{E\left(t_m\right)\right\}=\varsigma\left(B_i,t_m\right)$ is the joint probability of E(*tm*) on

Cubic_DG(*Bi,tm*). $\Pr\left\{H_{kj}E\left(t_m\right)\right\}$ is the joint probability of *Hkj*E(*tm*) on Cubic_DG(*Bi,tm*).

$\xi_i\left(B_i,t_m\right)$ is the weight factor of *E(tm)* on different Cubic_DG(*Bi,tm*)s calculated by

Equation (4), $\xi_i\left(B_i,t_m\right)$=1, when there is only one Cubic_DG(*Bi,tm*).

$$\Pr\left\{H_{kj}\left(t_m\right)\right\}=h_{kj}^s\left(t_m\right)=\xi_i\left(B_i,t_m\right)\frac{\Pr\left\{H_{kj}E\left(t_m\right)\right\}}{\Pr\left\{E\left(t_m\right)\right\}} \tag{3}$$

$$\xi_i\left(B_i,t_m\right)=\frac{\Pr\left\{E\left(t_m\right)\right\}}{\sum_i\Pr\left\{E\left(t_m\right)\right\}}=\frac{\varsigma\left(B_i,t_m\right)}{\sum_i\varsigma\left(B_i,t_m\right)} \tag{4}$$

Receive the evidence at the next time (*tm+1*), repeat the reasoning process of step 2 to 4 mentioned above, until the fault is diagnosed.

This is the inference calculation process of the cubic DUCG, and the inference is based on time series. The algorithm reconstructs the current cubic DUCG based on the evidence received at the current moment and the cubic DUCG at the previous moment, showing the causal propagation process based on time series. In order to explain the calculation process of cubic DUCG in detail, a case is given in Appendix A.

## 3. System Design

According to the reasoning mode of cubic DUCG and the characteristics of industrial diagnostic systems, the cubic DUCG based industrial fault diagnostic system is divided into four parts, communication module, real-time monitoring and diagnosis module, inference engine, and knowledge editing tool. The summary of each functional module is shown in Figure 4, the communication module is designed to process real-time data received from an industrial system. The knowledge editing tool is employed by domain experts to design DUCG knowledge bases. The inference engine is the algorithm implementation of the cubic DUCG, it is used for fault reasoning. The real-time monitoring and diagnosis module is the control center of the system, it monitors abnormal signals, sends the data, and the selected DUCG knowledge base together to the inference engine for fault diagnosis and demonstrates diagnostic results.

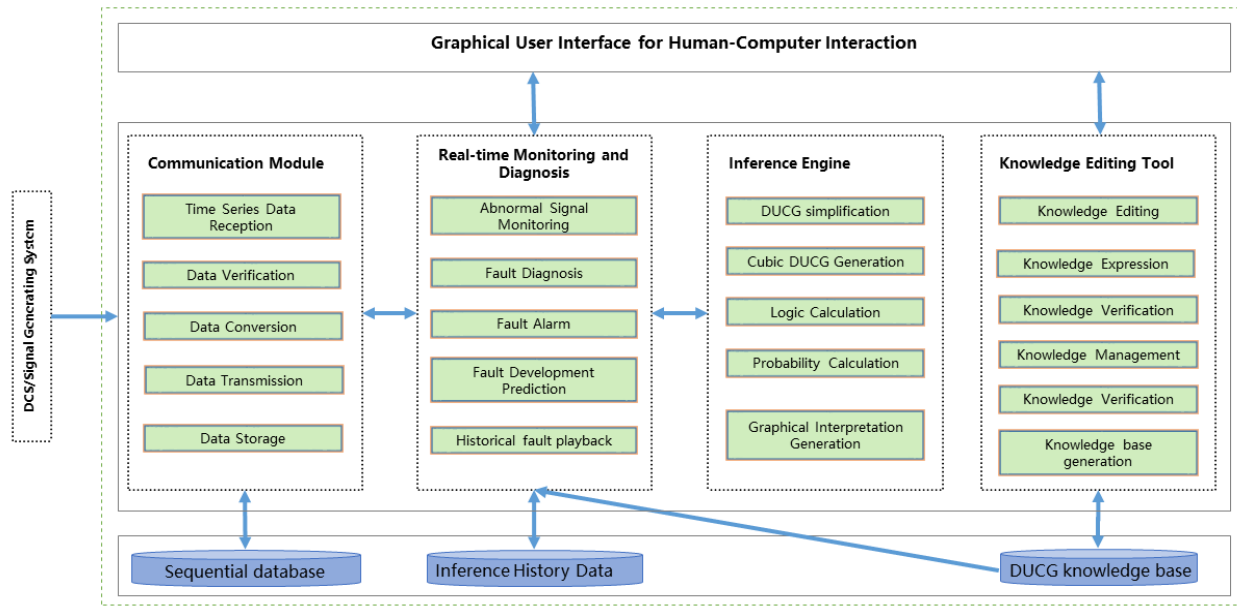

**Figure 4**. System block diagram of the cubic DUCG based industrial fault diagnostic system.

**Knowledge editing tool**: This module is used to design the DUCG knowledge base by domain experts with their experience and knowledge. The DUCG knowledge base can be built in a modular way, a whole DUCG can be divided into several sub-DUCGs. Generally, each sub-DUCG contains one fault, represents causal relations between the fault and its related monitoring signals, this modular knowledge construction method can reduce the difficulty of constructing a large complex knowledge base. The variables in sub-DUCG can be created in different manners, they can be imported from external files or drag the different icon in the toolbar on the right to create it directly, and all variables can be shared by different sub-DUCGs. Figure 5(a) is the sub-DUCG of condensate extraction pump failure, it describes the changes of its monitoring signals when condensate extraction pump failure. The square stands for condensate extraction pump failure($B_1$), the circles stand for different signals related to $B_1$, the variable can be edited by double-clicking its shape, the detailed information of a variable is shown in Figure 5(b). Causal relations between variables are represented by red directs, the intensity of the causal relationship can be edited by double-clicking the direct red line shown in Figure 5(c). In a sub-DUCG, when all the variables are created and all the relationships are expressed, then the sub-DUCG is finished. When sub-DUCGs are finished, the user can choose part or all sub-DUCGs to compile and generate the DUCG knowledge base. This process is performed automatically by the system. During compilation, some validations are performed to verify if the sub-DUCG conforms to rules of DUCG construction, including removing the redundant variables and relationships in different sub-DUCGs. After verification, the system automatically synthesizes a DUCG knowledge base that can be used for reasoning. Figure 6 shows a complete DUCG knowledge base, it is used for the fault diagnosis of the secondary circuit of unit 1 at the Ningde Nuclear Power Plant. This DUCG contains 24 $B$-type variables that stand for 24 different fault sources in the secondary circuit of nuclear voltage water reactor, the detail of the faults are shown in Appendix B. 141 $X$-type variables are used to describe the intermediate process or result arising from a root fault, a total of 1192 $F$-type variables (the direct red line) are used to describe the causal relations among variables.

**Figure 5**. The main UI of DUCG knowledge base edit. Here, users can create and edit variables, sub-DUCGs, compile and generate a DUCG knowledge base. (a) is the sub-DUCG of condensate extraction pump failure. (b) is the form of an X-type variable. It describes the basic information of a variable, including variable description, brief information, keywords, measure point, variable classification, and states of the variable. Among them, the measuring point is used to map the input signal to the variable, and the numerical interval of the state is used to map the value of the input signal to the state of a variable. The B-type variable is similar to the X-type variable, it has no numerical interval, but has a prior probability for each state. (c) is the interface for editing probability intensity. It is used to set the probability that a parent variable causes its child variable to occur. In this figure, it describes that when condensate pump CEX001PO is shut down due to failure, the probability that it will cause CEX002PO pump outlet pressure to decrease is 0.9.

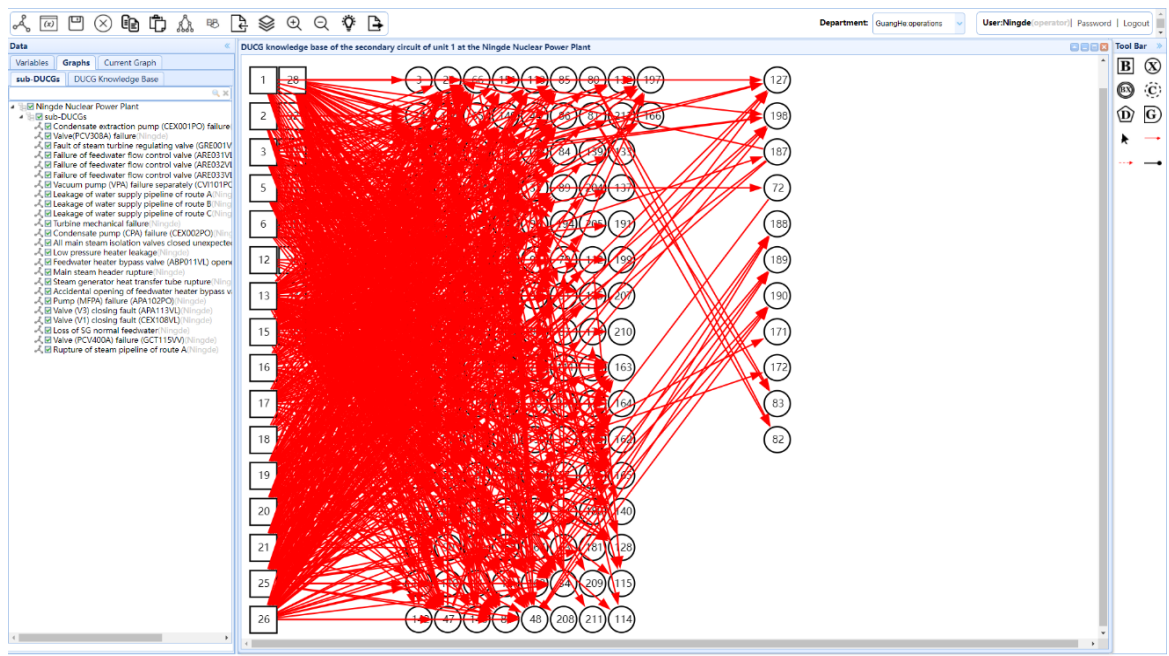

**Figure 6.** The DUCG knowledge base about the secondary circuit of unit 1 at Ningde nuclear power plant, contains 24 sub-DUCGs, each sub-DUCG stands for one root fault source.

**Communication module**: Its function is signal processing. The communication module receives the monitoring data from the industrial system, and transforms the data to make it conform to the data format requirements of DUCG according to the mapping relationship between measure points and variables, then transmits the data to real-time monitoring and diagnosis module.

**Inference engine**: The inference engine is the core module of the fault diagnostic system. It is based on the theory of cubic DUCG, which can perform the fault diagnosis and predict the subsequent development of fault. When it is used for fault diagnosis, it can generate cubic DUCG and do continuous causal reasoning based on the cubic DUCG at the moment that the abnormal evidence is received. Its diagnostic results are presented in probabilistic form, and the generated cubic DUCG is used to explain the results. When it performs fault development prediction, it can predict the possible abnormal conditions in the next time according to the current fault, abnormal signals, and DUCG knowledge base. The inference engine is an independent service. Its data comes from the real-time monitoring module, and its inference results will be sent back to the real-time monitoring module for user to decision making.

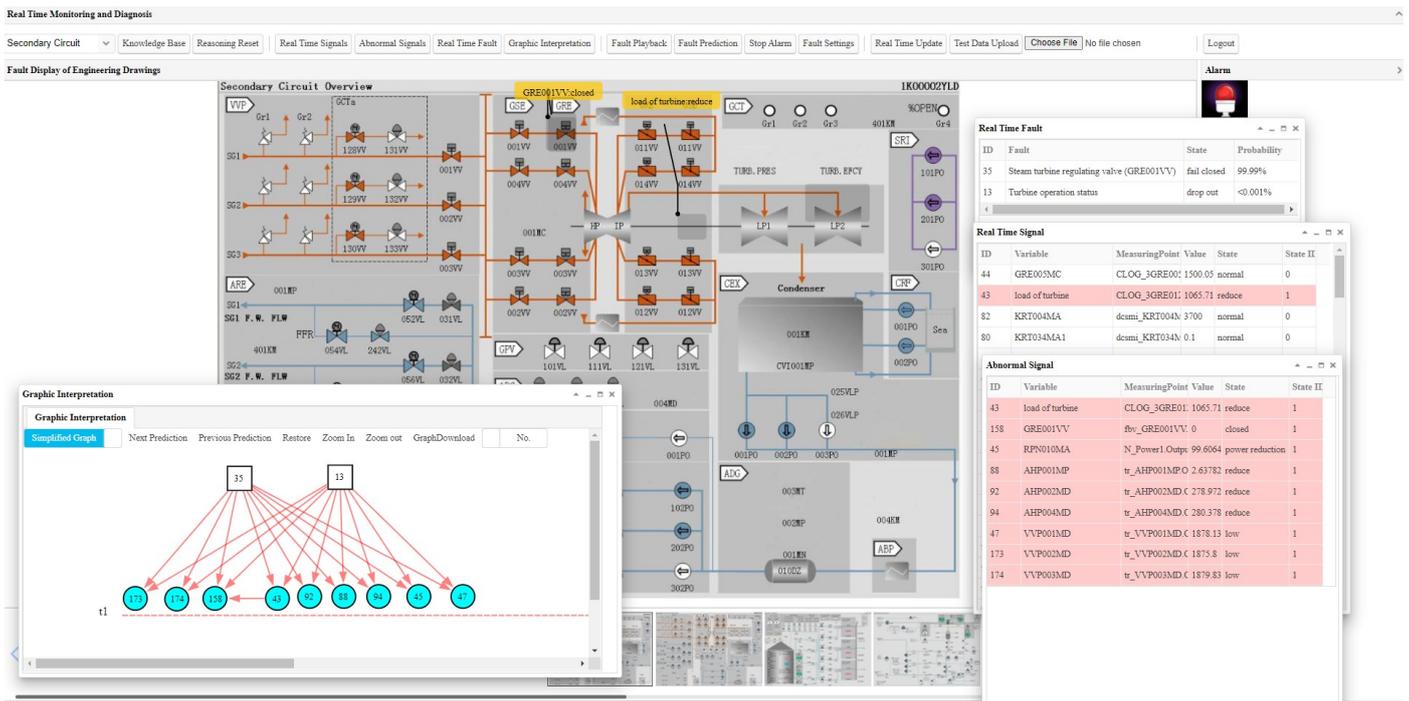

**Figure 7**. The main user interface of the real-time monitoring and diagnosis module.

**Real-time monitoring and diagnosis module:** The real-time monitoring and diagnosis module is the control and human-computer interaction center of the whole system. In this module, users can choose the DUCG knowledge base based on monitoring requirements, then the instruction is sent to the communication module to receive and process signals associated with the current DUCG knowledge base. The monitoring module displays and monitors signals in real-time. As shown in Figure 7, when abnormal signals appear, the monitoring module gives an alarm and sends signals to the inference engine for fault diagnosis, the diagnostic result and the graphical interpretation is sent back for display, meanwhile, the potential fault and abnormal signals will be displayed in the corresponding position of the engineering drawing.

The four functional modules of the system complete the functions of building a knowledge base, receiving and processing data, diagnosing faults, and displaying results. Through this system, users can translate knowledge and experience into diagnostic models. And use the model for real-time fault diagnosis.

## 4. Experiment

In order to validate the feasibility and diagnostic accuracy of the system, an experiment was made based on the secondary circuit of unit 1 at Ningde Power Plant, the DUCG knowledge base used in the test is shown in Figure 6. The system was deployed on the cluster. The inference engine is deployed on a machine with the CPU of AMD Ryzen 7 5700G@4.45GHz, 8-core processor, and 128GB RAM. Due to the extremely low accident rate of nuclear power plants, the data of the experiment was collected from the simulator of the secondary circuit of unit 1 at Ningde Power Plant. The fault was set on the simulator to generate the corresponding real-time fault data, which was sent to the communication program of the fault diagnosis system for the fault monitoring and diagnosis reasoning, and the data was sent every second. The nuclear power plant units in the simulator during the test were in normal operation state at full power at the beginning, and then the fault was inserted into the simulator. A total of 24 cases are used to test the system, each case verifies one fault. An example of condensate extraction pump failure (code: CEX001PO) is used to demonstrate the process of the experiment.

The fault "condensate extraction pump fault (CEX001PO)" was inserted at the 13th second after the simulator operated stably, and the opening of the pump CEX001PO gradually decreased, then, the communication module received the real-time data at 14th second, one of the variables was in its abnormal state (Intake pressure of ABP401RE is low(ABP004MP), $X_{71,2}$), and this moment was marked as $t_1$ in our diagnostic system. Due to the existence of the abnormal signal, the system was alarming, and this group of data was transmitted to the inference engine, the reasoning calculation results at $t_1$ as shown in Figure 8, and the graphic explanation is shown in Figure. 9.

| ID | Fault | State | Probability |
|---|---|---|---|
| | **Real Time Fault** | | |
| 36 | Condenser vacuum pump (CVI101PO) | failure | 61.07% |
| 1 | Condensate extraction pump status (CEX001PO) | closed | 35.87% |
| 2 | Condensate extraction pump status (CEX002PO) | closed | 3.05% |
| 26 | Water supply valve of electric main water supply system (APA113VL) | fail closed | <0.001% |
| 3 | Feedwater flow control valve status (ARE031VL) | all-open | <0.001% |
| 5 | Feedwater flow control valve status (ARE032VL) | all-open | <0.001% |
| 28 | Condensate valve (CEX108VL) | fail closed | <0.001% |
| 6 | Feedwater flow control valve status (ARE033VL) | all-open | <0.001% |
| 32 | Turbine bypass valve (GCT115VV) | fail opened | <0.001% |
| 33 | Status of A-way steam pipeline | rupture | <0.001% |
| 35 | Steam turbine regulating valve (GRE001VV) | fail opened | <0.001% |
| 25 | Pump (MFPA) failure (APA102PO) | fault | <0.001% |
| 38 | SG feedwater status | lose | <0.001% |
| 15 | Low pressure heater pipe (ABP401RE) | rupture | <0.001% |
| 16 | Feedwater heater bypass valve status (ABP011VL) | Unexpected open | <0.001% |
| 19 | Feedwater heater bypass valve (AHP009VL) | unexpected open | <0.001% |

**Figure 8.** The inference results at $t_1$ according to the abnormal evidence $E = X_{71,2}$

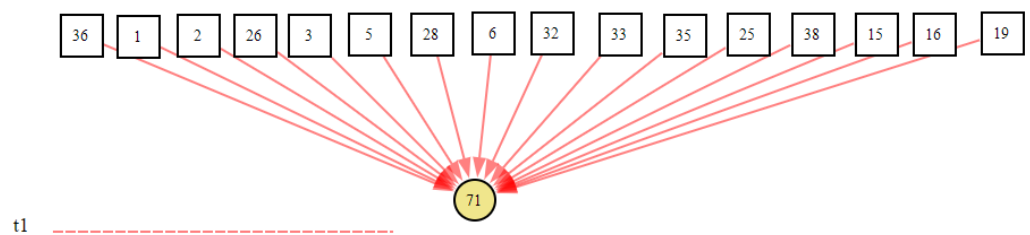

**Figure 9.** The graphic interpretation of inference hypothesis and abnormal evidence at $t_1$.

As we can see from the results, because multiple fault sources can cause $X_{71,2}$ to occur, the fault source cannot be diagnosed at $t_1$, but the scope of the fault and the probability of each fault can be preliminarily inferred. 16 fault sources can cause $X_{71,2}$ to occur, and the top 3 faults in the result list are more likely to cause it to occur.

At 15th second, no new abnormal was received, so the inference engine did not work. At the 16th second, the communication module received the new abnormal signal(Condensate extraction pump (CEX003PO) failure, $X_{195,1}$), this time was marked as $t_2$. Due to the emergence of new evidence, the data was sent to the inference engine for the second inference. The inference results are shown in Figure 10 and the graphic explanation is shown in Figure 11. Comparing the inference results in Figure 8 and the results in Figure 10, we can see that hypothesis space is further reduced. Because those hypotheses in first inference that cannot explain the new evidence were excluded, and only the hypotheses can explain all abnormal evidence were retained. According to the results, we can inference that the abnormal signals are possibly caused by condensate extraction pump status (CEX001PO)($B_{1,1}$) or condensate extraction pump status (CEX002PO)($B_{2,1}$), but $B_{1,1}$ is more likely.

**Real Time Fault**

| ID | Fault | State | Probability |
|----|-------|-------|-------------|
| 1 | Condensate extraction pump status (CEX001PO) | closed | 54.02% |
| 2 | Condensate extraction pump status (CEX002PO) | closed | 45.98% |
| 25 | Pump (MFPA) failure (APA102PO) | fault | <0.001% |

**Figure 10**. The inference results at $t_2$ according to the abnormal evidence $E=X_{71,2}X_{195,1}$

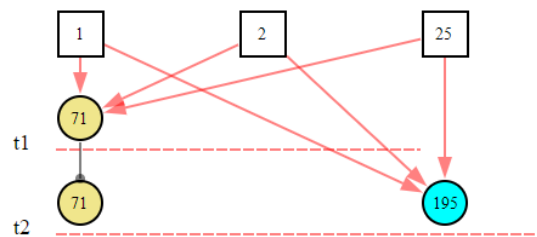

**Figure 11**. The graphic interpretation of inference hypothesis and abnormal evidence at $t_2$.

At the 17th second, the communication system received more abnormal signals. This time was marked as time $t_3$. According to the received new abnormal evidence, the third inference results and graphic interpretation were shown in Figure 12 and 13. Because only $B_{1,1}$ can explain all known abnormal evidence, according to the current evidence, $B_{1,1}$ was diagnosed.

**Real Time Fault**

| ID | Fault | State | Probability |
|----|-------|-------|-------------|
| 1 | Condensate extraction pump status (CEX001PO) | closed | 99.99% |

**Figure 12**. The thrid inference results when more abnormal evidence were received.

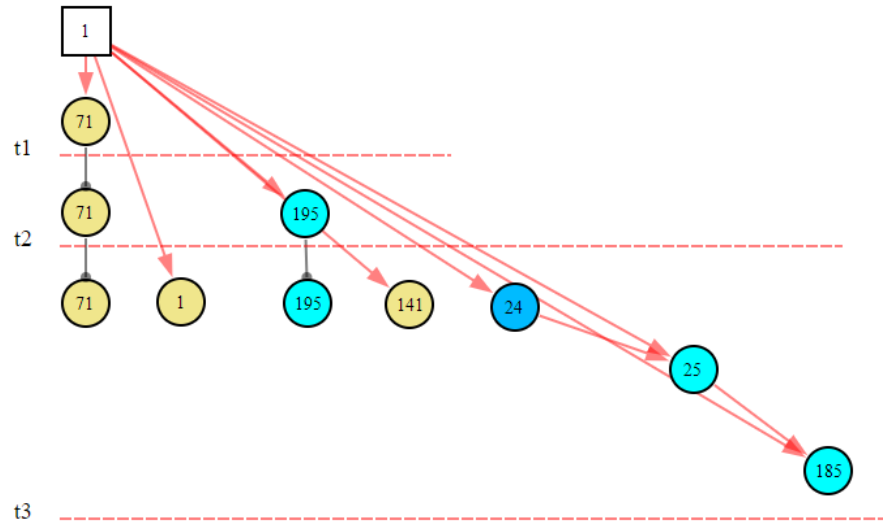

**Figure 13**. The graphic interpretation of inference hypothesis and abnormal evidence at *t₃*.

In the next few moments, the system continued to receive some new abnormal evidence, the inference engine continued to reason and calculate. The final graphic explanation was shown in Figure 14, $B_{1,1}$ can explain all abnormal evidence from beginning to end. Until *t₈*, no new abnormal evidence appeared and the diagnosis was over.

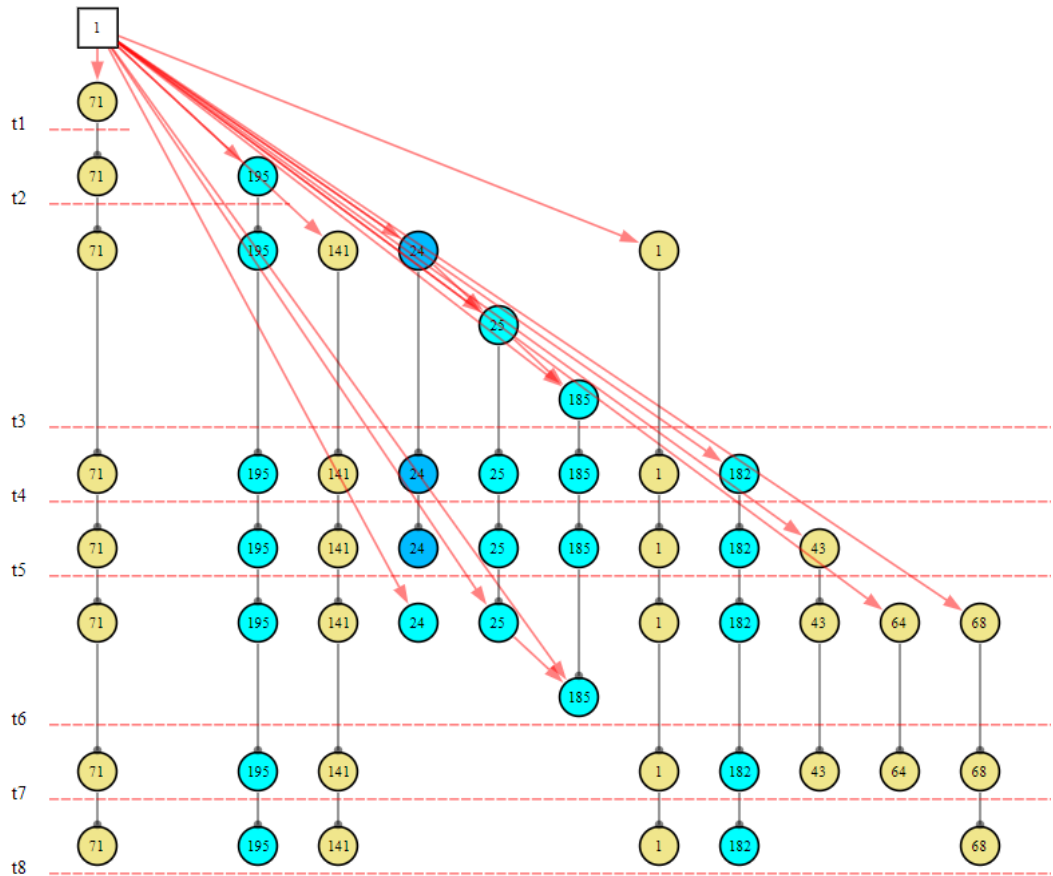

**Figure 14**. The graphic interpretation of inference hypothesis and abnormal evidence at *t₈*.

This example shows the diagnosis process of cubic-DUCG, It performs inference calculations based on time series. This fault diagnosis includes a total of 8 moments, the moment that first received the abnormal evidence is marked as $t_1$. At $t_1$, there is less abnormal evidence, the specific diagnosis result cannot be determined, but the range of possible failures can be roughly determined. At $t_2$, new abnormal evidence is added to the diagnosis, which further narrows the scope of the fault diagnosis. At $t_3$, new evidence increases, and the fault is uniquely determined. In the next few moments, new evidence continued to be received, but since there is only one diagnostic fault left, and the fault could explain all abnormal evidence, the diagnosis result does not change, and the diagnosis is completed. At each moment, the system dynamically generates a new cubic DUCG for diagnosis combined with the cubic DUCG obtained from the diagnosis at the previous moment and the new evidence, giving the reasoning diagnosis results. The graphic interpretations can clearly demonstrate the development and evolution process of the fault with time. It is convenient for the operator to understand the development of the fault for troubleshooting.

**Table 2**. Test results for 24 samples. Fault is the code of the fault in DUCG knowledge base. First Suspected indicates the starting moment when the fault is ranked first in the diagnostic results. Confirmed Diagnosis indicates the moment when the fault was confirmed. Time Consuming indicates the total reasoning time for diagnosing the fault.

| Fault | First Suspected | Confirmed Diagnosis | Time Consuming | Average Time |
|---|---|---|---|---|
| $B_{1,1}$ | $t_2$ | $t_3$ | 145ms | 48ms |
| $B_{2,1}$ | $t_1$ | $t_1$ | 19ms | 19ms |
| $B_{3,2}$ | $t_1$ | $t_1$ | 18ms | 18ms |
| $B_{5,2}$ | $t_1$ | $t_1$ | 19ms | 19ms |
| $B_{6,2}$ | $t_1$ | $t_1$ | 18ms | 18ms |
| $B_{13,1}$ | $t_1$ | $t_1$ | 313ms | 313ms |
| $B_{15,1}$ | $t_1$ | $t_1$ | 26ms | 26ms |
| $B_{16,1}$ | $t_3$ | $t_8$ | 437ms | 55ms |
| $B_{17,1}$ | $t_1$ | $t_1$ | 21ms | 21ms |
| $B_{18,1}$ | $t_1$ | $t_1$ | 20ms | 20ms |
| $B_{19,1}$ | $t_3$ | $t_8$ | 433ms | 55ms |
| $B_{20,1}$ | $t_1$ | $t_1$ | 274ms | 274ms |
| $B_{21,1}$ | $t_1$ | $t_1$ | 276ms | 276ms |
| $B_{25,1}$ | $t_1$ | $t_1$ | 81ms | 81ms |
| $B_{26,1}$ | $t_2$ | $t_2$ | 188ms | 94ms |
| $B_{28,1}$ | $t_1$ | $t_1$ | 23ms | 23ms |
| $B_{32,1}$ | $t_1$ | $t_1$ | 142ms | 142ms |
| $B_{33,1}$ | $t_1$ | $t_1$ | 230ms | 230ms |
| $B_{34,1}$ | $t_4$ | $t_4$ | 496ms | 124ms |
| $B_{35,1}$ | $t_1$ | $t_{12}$ | 1874ms | 157ms |
| $B_{35,2}$ | $t_1$ | $t_1$ | 31ms | 31ms |
| $B_{36,1}$ | $t_1$ | $t_1$ | 18ms | 18ms |
| $B_{37,1}$ | $t_1$ | $t_1$ | 21ms | 21ms |
| $B_{38,1}$ | $t_2$ | $t_2$ | 398ms | 199ms |

It can be seen from the table that all 24 test faults are correctly diagnosed, and 17 of them can be diagnosed at time $t_1$. The remaining 7 faults need multiple time slices to be diagnosed, but they can all be sorted to the first place in the list of diagnostic faults within 4 diagnostic time slices, which proves that the system has a high diagnostic accuracy rate. In terms of inference calculation time, although the inference calculation time increases with the increase of the diagnosis time slice, the average time (from the start of diagnosis to the fault confirmed) for each diagnosis is within 400ms. It shows that the system can complete real-time and efficient reasoning diagnosis and meet the real-time task requirements of industrial diagnosis.

## 5. Conclusions

Industrial systems such as nuclear power plants and spacecraft have high safety and almost no fault data is generated. The traditional fault diagnosis algorithm based on data modeling cannot realize the fault diagnosis of this kind of industrial system. This paper realizes the industrial fault diagnosis system based on the cubic DUCG theory. The system can carry out continuous fault diagnosis according to the time sequence, give the diagnosis results, and explain the results with graphics to assist users in decision-making. The diagnosis model of the system is constructed based on expert knowledge and experience and is independent of sample data. The problem of industrial system modeling and real-time fault diagnosis without sample data is successfully solved. In order to verify the feasibility of the system, this paper constructs the fault diagnosis model of the system with the reference of the second circuit of unit 1 of the Ningde nuclear power plant. The system is verified with fault cases, and 24 faults are correctly diagnosed. The experiment shows that the system is feasible and has high diagnostic accuracy. Next, we will improve in the following two aspects, how to ensure the accuracy of expert knowledge and how to continue to improve the efficiency of inference calculation to meet the construction and inference needs of more complex fault diagnosis models.

**Author Contributions:** Conceptualization, X.B.,H.N.,Z.Z. and Q.Z.; methodology, X.B.,H.N. and Q.Z.; software, X.B.,H.N. and Z.Z.; validation, X.B. and H.N.; formal analysis, Z.Z. and Q.Z.; investigation, X.B.,H.N. and Z.Z.; resources, Z.Z. and Q.Z.; writing—original draft preparation, X.B.; writing—review and editing, X.B. and Q.Z.; visualization, X.B. and H.N. All authors have read and agreed to the published version of the manuscript.

**Funding:** This research received no external funding.

**Data Availability Statement:** Not applicable.

**Acknowledgments:** The authors would like to thank all authors of previous papers for approving the use of their published research results in this paper.

**Conflicts of Interest:** The authors declare no conflict of interest.

## Appendix A

An example is employed to illustrate the inference process of the cubic DUCG. The original DUCG knowledge base is shown in Figure 15. It contains two root faults, $B_1$ (It has two states, state 0 is its normal state, state 1 is its abnormal state) and $B_2$ (It has three states, state 0 is the normal state, state 1 and state 2 are two different abnormal states), Other Variables($X_{3-7}$) are some intermediate process or results caused by $B_1$ or $B_2$, the strengths of the causal relationship between different variables are shown in Figure 3. The whole inference process contains three time series, $t_1$, $t_2$, and $t_3$. This example shows the dynamic calculation process of the inference engine in a continuous time series.

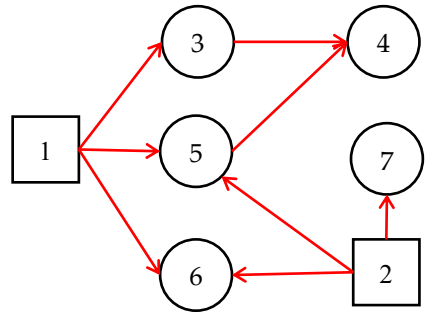

**Figure 15**. An example DUCG knowledge base used to demonstrate the inference process of the cubic DUCG.

**Table 3**. The paramaters of the DUCG knowledge base in Figure 15.

$$b_1 = \begin{pmatrix} - & 0.2 \end{pmatrix}^T, \ b_2 = \begin{pmatrix} - & 0.1 & 0.3 \end{pmatrix}^T, \ a_{3;1} = \begin{pmatrix} - & - \\ - & 0.5 \end{pmatrix}, \ a_{4;4} = \begin{pmatrix} - & - \\ - & 0.2 \end{pmatrix}, \ a_{5;1} = \begin{pmatrix} - & - \\ - & 0.1 \end{pmatrix}, \ a_{4;5} = \begin{pmatrix} - & - \\ - & 0.7 \end{pmatrix},$$

$$a_{6;1} = \begin{pmatrix} - & - \\ - & 0.4 \end{pmatrix}, a_{6;2} = \begin{pmatrix} - & - & - \\ - & 0.5 & 0.9 \end{pmatrix}, a_{5;2} = \begin{pmatrix} - & - & - \\ - & 0.5 & 0.5 \end{pmatrix}, a_{7;2} = \begin{pmatrix} - & - & - \\ - & 0.3 & 0.8 \end{pmatrix}, \ r_i = 1.$$

**Inference process**:

**Step 1.1**. DUCG decomposition. We obtain two sub-DUCGs, sub-DUCG($B_1$) and sub-DUCG($B_2$) by decomposing the original DUCG in Figure 15. Each sub-DUCG($B_i$) describe the relationships between the fault $B_i$ and its related variables, the sub-DUCGs are shown in Figure 16.

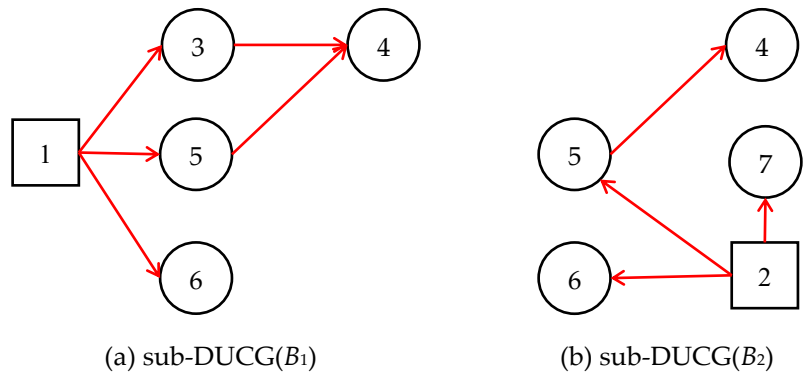

(a) sub-DUCG($B_1$)          (b) sub-DUCG($B_2$)

**Figure 16**. The original DUCG are decomposing into two sub-DUCG($B_i$)s.

**Step 1.2**. DUCG($B_i$) simplification. As shown in Figure 17, suppose the evidence at $t_1$ is $E(t_1) = X_{3,0}X_{5,1}X_{6,0}$. According to the simplification rules of DUCG, the unrelated variables and relations are removed, we get two Slice_DG($B_i$)s at $t_1$ by simplifing sub-DUCG($B_1$) and sub-DUCG($B_2$), the results are shown in Figure 18(a) and (b).

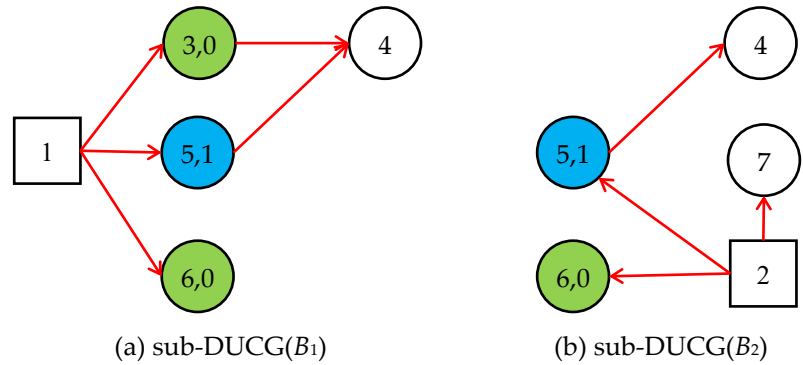

(a) sub-DUCG($B_1$)  (b) sub-DUCG($B_2$)

**Figure 17**. The status of two sub-DUCGs at $t_1$.

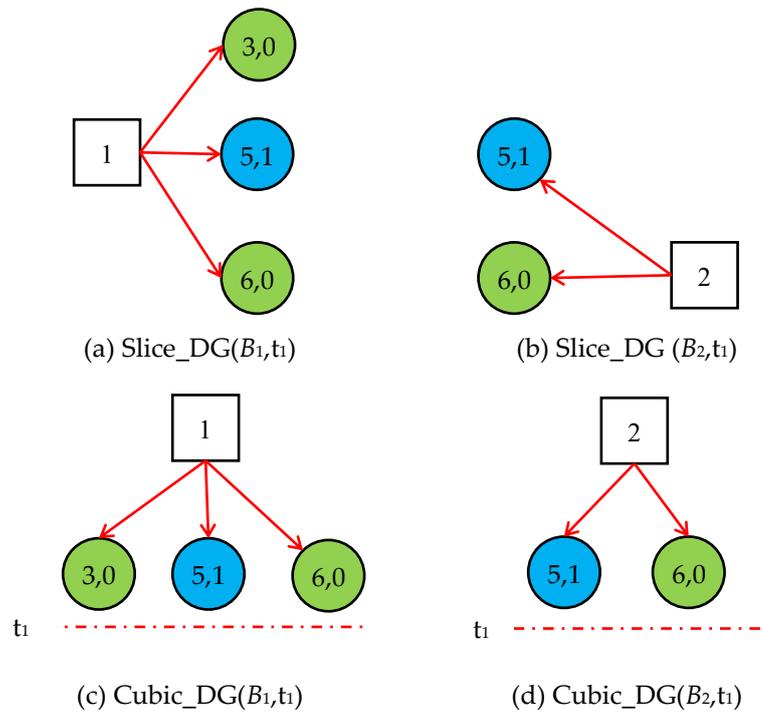

(a) Slice_DG($B_1$,$t_1$)  (b) Slice_DG ($B_2$,$t_1$)

(c) Cubic_DG($B_1$,$t_1$)  (d) Cubic_DG($B_2$,$t_1$)

**Figure 18**. The Slice_DG($B_i$,$t_m$)s and Cubic_DB($B_i$,$t_m$)s at $t_1$. (a) Slice_DG($B_1$,$t_1$) is obtained by simplifying the sub-DUCG($B_1$) under the evidence $E$($t_1$). (b) is obtained by simplifying the sub-DUCG($B_2$). (c) is the Cubic_DG($B_1$,$t_1$), it is as same as Slice_DG($B_1$,$t_1$) at $t_1$. (d) is the Cubic_DG($B_2$,$t_1$).

**Step 1.3**. *Cubic_DB($B_i$,$t_m$)* generation. At $t_1$, the Slice_DG($B_i$,$t_1$)s are regarded as Cubic_DG($B_i$,$t_1$)s, as shown in Figure 18(c) and (d). From the two Cubic_DG($B_i$,$t_1$)s, we get the hypothesis spaces $S_H(t_1) = \{H_{1,1}, H_{2,1}, H_{2,1}\} = \{B_{1,1}, B_{2,1}, B_{2,2}\}$.

**Step 1.4**. Probability reasoning. The hypothesis $H_{1,1}$ is included in Cubic_DG($B_1$,$t_1$), and The pypothesises $H_{2,1}$ and $H_{2,2}$ are inclueded in Cubic_DG($B_2$,$t_1$). In order to calculate the $\Pr\{H_{1,1}(t_1)\}$, we should do expression expansion of $E(t_1)$ and $H_{1,1}E(t_1)$ based on Cubic_DG($B_1,t_1$), and then, calculate $\Pr\{E(t_1)\}$, $\Pr\{H_{1,1}E(t_1)\}$, the results are shown in Equation (5-8) respectively.

$$E\left(t_1\right) = X_{3,0}X_{5,1}X_{6,0} = F_{3,0;1,1}B_{1,1}F_{5,1;1,1}B_{1,1}F_{6,0;1,1}B_{1,1} = F_{3,0;1,1}F_{5,1;1,1}F_{6,0;1,1}B_{1,1} \qquad (5)$$

$$H_{1,1}E\left(t_1\right) = B_{1,1}X_{3,0}X_{5,1}X_{6,0} = F_{3,0;1,1}F_{5,1;1,1}F_{6,0;1,1}B_{1,1} \qquad (6)$$

$$\begin{aligned}
\varsigma\left(B_1,t_1\right) &= \Pr\left\{E\left(t_1\right)\right\} = \Pr\left\{X_{3,0}X_{5,1}X_{6,0}\right\} = \Pr\left\{F_{3,0;1,1}F_{5,1;1,1}F_{6,0;1,1}B_{1,1}\right\} \\
&= f_{3,0;1,1}f_{5,1;1,1}f_{6,0;1,1}b_{1,1} \\
&= \left(r_1/r_3\right)a_{3,0;1,1}\left(r_1/r_5\right)a_{5,1;1,1}\left(r_1/r_6\right)a_{6,0;1,1}b_{1,1} \\
&\left(r_1/r_3\right)\times\left(1-a_{3,1;1,1}\right)\times\left(r_1/r_5\right)\times a_{5,1;1,1}\times\left(r_1/r_6\right)\times\left(1-a_{6,1;1,1}\right)\times b_{1,1} \\
&= \left(1/1\right)\times\left(1-0.5\right)\times\left(1/1\right)\times 0.1\times\left(1/1\right)\times\left(1-0.4\right)\times 0.2 \\
&= 0.006
\end{aligned} \qquad (7)$$

$$\Pr\left\{H_{1,1}E\left(t_1\right)\right\} = \Pr\left\{B_{1,1}X_{3,0}X_{5,1}X_{6,0}\right\} = \Pr\left\{F_{3,0;1,1}F_{5,1;1,1}F_{6,0;1,1}B_{1,1}\right\} = 0.006 \qquad (8)$$

Similarly, for $H_{2,1}$ and $H_{2,2}$ in Cubic_DG($B_2,t_1$). The results of expression expansion for $E(t_1)$ and $H_{1,1}E(t_1)$, and the results of $\Pr\{E(t_1)\}$, $\Pr\{H_{1,1}E(t_1)\}$, $\Pr\{H_{2,1}(t_1)\}$ are shown in Equation (9-13).

$$\begin{aligned}
E\left(t_1\right) &= X_{5,1}X_{6,0} \\
&= F_{5,1;2}B_2F_{6,0;2}B_2 \\
&= \left(F_{5,1;2,1}B_{2,1}+F_{5,1;2,2}B_{2,2}\right)\left(F_{6,0;2,1}B_{2,1}+F_{6,0;2,2}B_{2,2}\right) \\
&= F_{5,1;2,1}B_{2,1}F_{6,0;2,1}B_{2,1}+F_{5,1;2,2}B_{2,2}F_{6,0;2,2}B_{2,2} \\
&= F_{5,1;2,1}F_{6,0;2,1}B_{2,1}+F_{5,1;2,2}F_{6,0;2,2}B_{2,2}
\end{aligned} \qquad (9)$$

$$\begin{aligned}
H_{2,1}E\left(t_1\right) &= B_{2,1}X_{5,1}X_{6,0} \\
&= \left(F_{5,1;2,1}F_{6,0;2,1}B_{2,1}+F_{5,1;2,2}F_{6,0;2,2}B_{2,2}\right)B_{2,1} \\
&= F_{5,1;2,1}F_{6,0;2,1}B_{2,1}
\end{aligned} \qquad (10)$$

$$\begin{aligned}
\varsigma\left(B_2,t_1\right) &= \Pr\left\{E\left(t_1\right)\right\} = \Pr\left\{F_{5,1;2,1}F_{6,0;2,1}B_{2,1}+F_{5,1;2,2}F_{6,0;2,2}B_{2,2}\right\} \\
&= f_{5,1;2,1}f_{6,0;2,1}b_{2,1}+f_{5,1;2,2}f_{6,0;2,2}b_{2,2} \\
&= \left(r_2/r_5\right)a_{5,1;2,1}\left(r_2/r_6\right)a_{6,0;2,1}b_{2,1}+\left(r_2/r_5\right)a_{5,1;2,2}\left(r_2/r_6\right)a_{6,0;2,2}b_{2,2} \\
&= \left(1/1\right)\times 0.5\times\left(1/1\right)\times\left(1-0.5\right)\times 0.1+\left(1/1\right)\times 0.5\times\left(1/1\right)\times\left(1-0.9\right)\times 0.3 \\
&= 0.025+0.015 = 0.04
\end{aligned} \qquad (11)$$

$$\begin{aligned}
\Pr\left\{H_{2,1}E\left(t_1\right)\right\} &= \Pr\left\{F_{5,1;2,1}F_{6,0;2,1}B_{2,1}\right\} \\
&= f_{5,1;2,1}f_{6,0;2,1}b_{2,1} \\
&= \left(r_2/r_5\right)a_{5,1;2,1}\left(r_2/r_6\right)a_{6,0;2,1}b_{2,1} \\
&= (1/1)\times0.5\times(1/1)\times(1-0.5)\times0.1 \\
&= 0.025
\end{aligned} \tag{12}$$

$$\Pr\left\{H_{2,2}E\left(t_1\right)\right\} = \Pr\left\{F_{5,1;2,2}F_{6,0;2,2}B_{2,2}\right\} = 0.015 \tag{13}$$

According to Equation (3) in this paper, we get the posterior probability of $H_{1,1}$, $H_{2,1}$ and $H_{2,2}$ shown in Equation (14-16):

$$\begin{aligned}
\Pr\left\{H_{1,1}\left(t_1\right)\right\} &= h_{1,1}^s\left(t_1\right) = \xi_1\left(B_1,t_1\right)\frac{\Pr\left\{H_{1,1}E\left(t_1\right)\right\}}{\Pr\left\{E\left(t_1\right)\right\}} \\
&= \frac{\varsigma\left(B_1,t_1\right)}{\varsigma\left(B_1,t_1\right)+\varsigma\left(B_2,t_1\right)}\times\frac{\Pr\left\{H_{1,1}E\left(t_1\right)\right\}}{\Pr\left\{E\left(t_1\right)\right\}} \\
&= \frac{0.006}{0.006+0.04}\times\frac{0.006}{0.006} \\
&= 0.13
\end{aligned} \tag{14}$$

$$\begin{aligned}
\Pr\left\{H_{2,1}\left(t_1\right)\right\} &= h_{2,1}^s\left(t_1\right) = \xi_1\left(B_2,t_1\right)\frac{\Pr\left\{H_{2,1}E\left(t_1\right)\right\}}{\Pr\left\{E\left(t_1\right)\right\}} \\
&= \frac{\varsigma\left(B_2,t_1\right)}{\varsigma\left(B_1,t_1\right)+\varsigma\left(B_2,t_1\right)}\times\frac{\Pr\left\{H_{2,1}E\left(t_1\right)\right\}}{\Pr\left\{E\left(t_1\right)\right\}} \\
&= \frac{0.04}{0.006+0.04}\times\frac{0.025}{0.04} \\
&= 0.543
\end{aligned} \tag{15}$$

$$\begin{aligned}
\Pr\left\{H_{2,2}\left(t_1\right)\right\} &= h_{2,2}^s\left(t_1\right) = \xi_1\left(B_2,t_1\right)\frac{\Pr\left\{H_{2,2}E\left(t_1\right)\right\}}{\Pr\left\{E\left(t_1\right)\right\}} \\
&= \frac{\varsigma\left(B_2,t_1\right)}{\varsigma\left(B_1,t_1\right)+\varsigma\left(B_2,t_1\right)}\times\frac{\Pr\left\{H_{2,2}E\left(t_1\right)\right\}}{\Pr\left\{E\left(t_1\right)\right\}} \\
&= 0.326
\end{aligned} \tag{16}$$

From the results, we can see that $H_{2,1}$ are more likely to cause the occurrence of $E=X_{3,0}X_{5,1}X_{6,0}$ at $t_1$.

At $t_2$, suppose there is a new abnormal evidence $X_{6,1}$ occur, combined with the evidence at $t_1$, the total evidence at $t_2$ is $E(t_2)= X_{3,0}X_{5,1}X_{6,1}$, because new abnormal is received, then the new inference will begin.

**Step 2.1**. Simplified the original sub-DUCG($B_i$)s in Figure 16 under the evidence $E(t_2)$, we get the Slice_DG($B_1,t_2$) and Slice_DG($B_2,t_2$) shown in Figure 19.

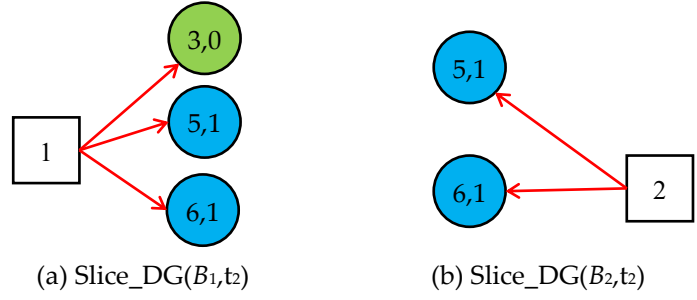

(a) Slice_DG($B_1$,t$_2$)  (b) Slice_DG($B_2$,t$_2$)

**Figure 19.** Slice_DG($B_i$,t$_2$) at t$_2$.

**Step 2.2.** By synthesizing the Cubic_DG($B_i$,t$_1$)s at t$_1$ and the Slice_DG($B_i$,t$_2$)s at t$_2$, we get the Cubic_DG($B_i$,t$_2$)s shown in Figure 20. Form Cubic_DG($B_i$,t$_2$)s, we can see that the states of $X_3$ and $X_5$ did not change, but the state of $X6$ changed from normal to abnormal.

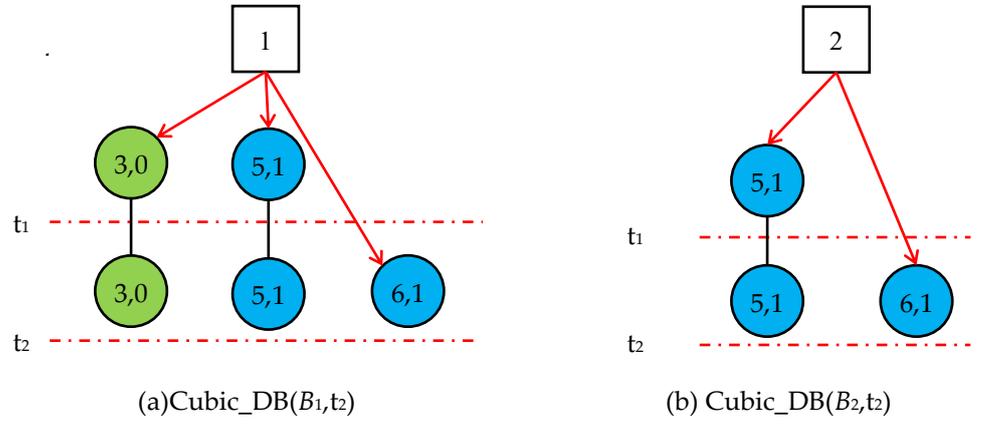

(a)Cubic_DB($B_1$,t$_2$)  (b) Cubic_DB($B_2$,t$_2$)

**Figure 20.** Cubic_DG($B_i$,t$_2$)s at t$_2$.

**Step 2.3.** At t$_2$, due to the emergence of new evidence $X_{6,1}$, Cubic_DB($B_i$,t$_2$)s are regenerated, so the posterior probability of each hypothesis needs to be recalculated under the evidence $E$(t$_2$) based on Cubic_DB($B_i$,t$_2$)s. For $H_{1,1}$ in Cubic_DB($B_1$,t$_2$).

$$E\left(t_2\right) = X_{3,0}X_{5,1}X_{6,1} = F_{3,0;1,1}B_{1,1}F_{5,1;1,1}B_{1,1}F_{6,1;1,1}B_{1,1} = F_{3,0;1,1}F_{5,1;1,1}F_{6,1;1,1}B_{1,1} \tag{17}$$

$$\begin{aligned}
\varsigma\left(B_1,t_2\right) &= \Pr\left\{E\left(t_2\right)\right\} = \Pr\left\{F_{3,0;1,1}F_{5,1;1,1}F_{6,1;1,1}B_{1,1}\right\} = f_{3,0;1,1}f_{5,1;1,1}f_{6,1;1,1}b_{1,1} \\
&= \left(r_1/r_3\right)a_{3,0;1,1}\left(r_1/r_5\right)a_{5,1;1,1}\left(r_1/r_6\right)a_{6,1;1,1}b_{1,1} \\
&= 0.004
\end{aligned} \tag{18}$$

$$H_{1,1}E\left(t_2\right) = B_{1,1}X_{3,0}X_{5,1}X_{6,1} = F_{3,0;1,1}F_{5,1;1,1}F_{6,1;1,1}B_{1,1} \tag{19}$$

$$\Pr\left\{H_{1,1}E\left(t_2\right)\right\} = \Pr\left\{B_{1,1}X_{3,0}X_{5,1}X_{6,1}\right\} = \Pr\left\{F_{3,0;1,1}F_{5,1;1,1}F_{6,1;1,1}B_{1,1}\right\} = 0.004 \tag{20}$$

For $H_{2,1}$ and $H_{2,2}$ in Cubic_DB($B_2$,t$_2$).

$$E(t_2) = X_{5,1}X_{6,1}$$
$$= F_{5,1;2}B_2F_{6,1;2}B_2 \tag{21}$$
$$= F_{5,1;2,1}F_{6,1;2,1}B_{2,1} + F_{5,1;2,2}F_{6,1;2,2}B_{2,2}$$

$$\varsigma(B_2, t_2) = \Pr\{E(t_2)\} = \Pr\{F_{5,1;2,1}F_{6,1;2,1}B_{2,1} + F_{5,1;2,2}F_{6,1;2,2}B_{2,2}\}$$
$$= f_{5,1;2,1}f_{6,1;2,1}b_{2,1} + f_{5,1;2,2}f_{6,1;2,2}b_{2,2} \tag{22}$$
$$= 0.16$$

$$H_{2,1}E(t_2) = B_{2,1}X_{5,1}X_{6,1}$$
$$= \left(F_{5,1;2,1}F_{6,1;2,1}B_{2,1} + F_{5,1;2,2}F_{6,1;2,2}B_{2,2}\right)B_{2,1} \tag{23}$$
$$= F_{5,1;2,1}F_{6,1;2,1}B_{2,1}$$

$$\Pr\{H_{2,1}E(t_2)\} = 0.025 \tag{24}$$

$$\Pr\{H_{2,2}E(t_2)\} = 0.135 \tag{25}$$

Calculate the posterior probability of each hypothesis according to Equation (3), and we get:

$$\Pr\{H_{1,1}(t_2)\} = h_{1,1}^s(t_2) = \xi_1(B_1, t_2)\frac{\Pr\{H_{1,1}E(t_2)\}}{\Pr\{E(t_2)\}}$$
$$= \frac{\varsigma(B_1, t_2)}{\varsigma(B_1, t_2) + \varsigma(B_2, t_2)} \times \frac{\Pr\{H_{1,1}E(t_2)\}}{\Pr\{E(t_2)\}} \tag{26}$$
$$= \frac{0.004}{0.004 + 0.16} \times \frac{0.004}{0.004}$$
$$= 0.024$$

$$\Pr\{H_{2,1}(t_2)\} = h_{2,1}^s(t_2) = \xi_1(B_2, t_2)\frac{\Pr\{H_{2,1}E(t_2)\}}{\Pr\{E(t_2)\}}$$
$$= \frac{\varsigma(B_2, t_2)}{\varsigma(B_1, t_2) + \varsigma(B_2, t_2)} \times \frac{\Pr\{H_{2,1}E(t_2)\}}{\Pr\{E(t_2)\}} \tag{27}$$
$$= 0.152$$

$$\Pr\{H_{2,2}(t_2)\} = h_{2,2}^s(t_2) = \xi_2(B_2, t_2)\frac{\Pr\{H_{2,2}E(t_2)\}}{\Pr\{E(t_2)\}}$$
$$= \frac{\varsigma(B_2, t_2)}{\varsigma(B_1, t_2) + \varsigma(B_2, t_2)} \times \frac{\Pr\{H_{2,2}E(t_2)\}}{\Pr\{E(t_2)\}} \tag{28}$$
$$= \frac{0.16}{0.004 + 0.16} \times \frac{0.135}{0.16}$$
$$= 0.823$$

From the result, we can see that $H_{2.2}$ is more likely to cause this evidence to occur at $t_2$, the probability of $H_{1.1}$ and $H_{2.1}$ decreases.

**Step 3.1.** At $t_3$, some new abnormal evidence is received, simplifying the original sub-DUCG($B_i$) under $E(t_3)$, then, we get two Slice_DG($B_i$,$t_3$)s. In Slice_DG($B_1$,$t_3$), the abnormal variable $X_{7.1}$ is an isolated variable. $H_{1.1}$ cannot explain why $X_{7.1}$ occurs, so Slice_DG($B_1$,$t_3$) is regarded as an invalided Slice_DG, and deleted. Only Slice_DG($B_2$,$t_3$) is retained.

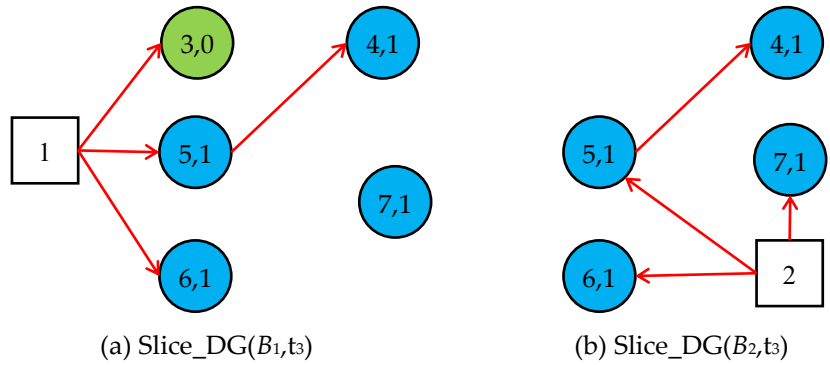

(a) Slice_DG($B_1$,$t_3$)      (b) Slice_DG($B_2$,$t_3$)

**Figure 21.** Slice_DG($B_i$,$t_3$) at $t_3$.

**Step 3.2.** According to Cubic_DG($B_2$,$t_2$) and Slice_DG($B_2$,$t_3$) to generate the Cubic_DG($B_2$,$t_3$) shown in Figure 22.

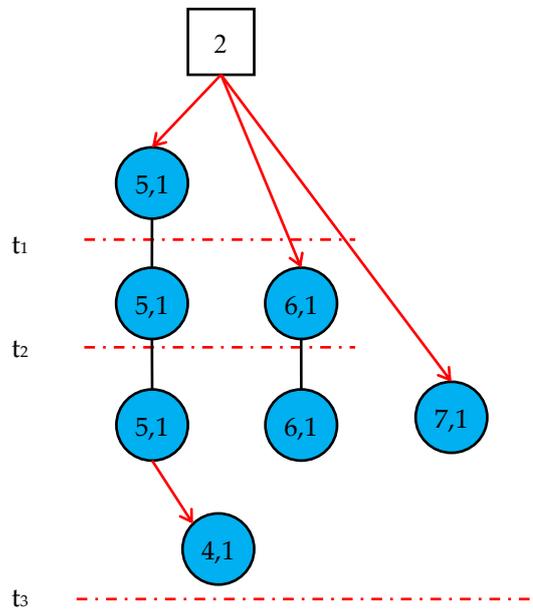

**Figure 22.** Cubic_DB($B_2$,$t_3$) at $t_3$.

**Step 3.3.** expend $E(t_3)$, $H_{2.1}E(t_3)$, $H_{2.2}E(t_3)$ based on Cubic_DB($B_2$,$t_3$) and calculate the joint probability of $\Pr\{E(t_3)\}$, $\Pr\{H_{2.1}E(t_3)\}$, $\Pr\{H_{2.2}E(t_3)\}$ shown in Equation (29-34).

$$
\begin{aligned}
E(t_3) &= X_{4,1}X_{5,1}X_{6,1}X_{7,1} \\
&= F_{4,1;5,1}F_{5,1;2}B_2F_{5,1;2}B_2F_{6,1;2}B_2F_{7,1;2}B_2 \\
&= F_{4,1;5,1}\left(F_{5,1;2,1}B_{2,1}+F_{5,1;2,2}B_{2,2}\right)\left(F_{6,1;2,1}B_{2,1}+F_{6,1;2,2}B_{2,2}\right)\left(F_{7,1;2,1}B_{2,1}+F_{7,1;2,2}B_{2,2}\right) \\
&= \left(F_{4,1;5,1}F_{5,1;2,1}B_{2,1}+F_{4,1;5,1}F_{5,1;2,2}B_{2,2}\right)\left(F_{6,1;2,1}B_{2,1}+F_{6,1;2,2}B_{2,2}\right)\left(F_{7,1;2,1}B_{2,1}+F_{7,1;2,2}B_{2,2}\right) \\
&= F_{4,1;5,1}F_{5,1;2,1}B_{2,1}F_{6,1;2,1}B_{2,1}F_{7,1;2,1}B_{2,1}+F_{4,1;5,1}F_{5,1;2,2}B_{2,2}F_{6,1;2,2}B_{2,2}F_{7,1;2,2}B_{2,2} \\
&= F_{4,1;5,1}F_{5,1;2,1}F_{6,1;2,1}F_{7,1;2,1}B_{2,1}+F_{4,1;5,1}F_{5,1;2,2}F_{6,1;2,2}F_{7,1;2,2}B_{2,2}
\end{aligned} \tag{29}
$$

$$
\begin{aligned}
\Pr\{E(t_3)\} &= \Pr\{F_{4,1;5,1}F_{5,1;2,1}F_{6,1;2,1}F_{7,1;2,1}B_{2,1}+F_{4,1;5,1}F_{5,1;2,2}F_{6,1;2,2}F_{7,1;2,2}B_{2,2}\} \\
&= f_{4,1;5,1}f_{5,1;2,1}f_{6,1;2,1}f_{7,1;2,1}b_{2,1}+f_{4,1;5,1}f_{5,1;2,2}f_{6,1;2,2}f_{7,1;2,2}b_{2,2} \\
&= 0.7\times0.5\times0.5\times0.3\times0.1+0.7\times0.5\times0.9\times0.8\times0.3 \\
&= 0.00525+0.0756 \\
&= 0.08085
\end{aligned} \tag{30}
$$

$$
H_{2,1}E(t_3) = X_{4,1}X_{5,1}X_{6,1}X_{7,1}B_{1,1} = F_{4,1;5,1}F_{5,1;2,1}F_{6,1;2,1}F_{7,1;2,1}B_{2,1} \tag{31}
$$

$$
\Pr\{H_{2,1}E(t_3)\} = \Pr\{F_{4,1;5,1}F_{5,1;2,1}F_{6,1;2,1}F_{7,1;2,1}B_{2,1}\} = 0.00525 \tag{32}
$$

$$
H_{2,2}E(t_3) = X_{4,1}X_{5,1}X_{6,1}X_{7,1}B_{2,2} = F_{4,1;5,1}F_{5,1;2,2}F_{6,1;2,2}F_{7,1;2,2}B_{2,2} \tag{33}
$$

$$
\Pr\{H_{2,2}E(t_3)\} = \Pr\{F_{4,1;5,1}F_{5,1;2,2}F_{6,1;2,2}F_{7,1;2,2}B_{2,2}\} = 0.0756 \tag{34}
$$

Because there is only one Slice_DG at $t_3$, so $\xi_2(B_2,t_3)=1$. According to Equation (3), we get the inference results of $H_{2,1}$ and $H_{2,2}$ shown in Equation (35) and (36).

$$
\Pr\{H_{2,1}(t_3)\} = \frac{\Pr\{H_{2,1}E(t_3)\}}{\Pr\{E(t_3)\}} = \frac{0.00525}{0.08085} = 0.065 \tag{35}
$$

$$
\Pr\{H_{2,2}(t_3)\} = \frac{\Pr\{H_{2,2}E(t_3)\}}{\Pr\{E(t_3)\}} = \frac{0.0756}{0.08085} = 0.935 \tag{36}
$$

The inference ended because there was no new evidence. From the three diagnostic processes, it can be seen that due to different evidence, the diagnostic results change with time, and finally get the correct diagnostic results.

### Appendix B

Description of 24 faults in the knowledge base, including fault name, state description, and its number in DUCG.

| ID | Fault Description | State | State Description |
|---|---|---|---|
| $B_1$ | Condensate extraction pump status (CEX001PO) | 0 | working order |
| | | 1 | closed |

| $B_2$ | Condensate extraction pump status (CEX002PO) | 0 | working order |
| | | 1 | closed |
| $B_3$ | Feedwater flow control valve status (ARE031VL) | 0 | normal |
| | | 1 | all closed |
| | | 2 | all opened |
| $B_5$ | Feedwater flow control valve status (ARE032VL) | 0 | normal |
| | | 1 | all closed |
| | | 2 | all opened |
| $B_6$ | Feedwater flow control valve status (ARE033VL) | 0 | normal |
| | | 1 | all closed |
| | | 2 | all opened |
| $B_{12}$ | Water supply pipeline of route A | 0 | normal |
| | | 1 | leaked |
| $B_{13}$ | Turbine operation status | 0 | working order |
| | | 1 | tripping state |
| $B_{15}$ | Low pressure heater pipe (ABP401RE) | 0 | normal |
| | | 1 | rupture |
| $B_{16}$ | Feedwater heater bypass valve status (ABP011VL) | 0 | normal |
| | | 1 | accidental opening |
| $B_{17}$ | Main steam header status | 0 | normal |
| | | 1 | leaked |
| $B_{18}$ | Steam generator heat transfer tube | 0 | normal |
| | | 1 | leaked |
| $B_{19}$ | Feedwater heater bypass valve (AHP009VL) | 0 | normal |
| | | 1 | accidental opening |
| $B_{20}$ | Water supply pipeline of route B | 0 | normal |
| | | 1 | leaked |
| $B_{21}$ | Water supply pipeline of route C | 0 | normal |
| | | 1 | leaked |
| $B_{25}$ | Pump (MFPA) failure (APA102PO) | 0 | normal |
| | | 1 | fault condition |
| $B_{26}$ | Water supply valve of electric main water supply system (APA113VL) | 0 | normal |
| | | 1 | accidental shutdown |
| $B_{28}$ | Water supply valve of electric main water supply system (APA113VL) | 0 | normal |
| | | 1 | accidental shutdown |
| $B_{32}$ | Turbine bypass valve (GCT115VV) | 0 | normal |
| | | 1 | accidental opening |
| $B_{33}$ | Status of A-way steam pipeline | 0 | normal |
| | | 1 | rupture |
| $B_{34}$ | Turbine bypass valve (GCT131VV) | 0 | normal |
| | | 1 | accidental opening |
| $B_{35}$ | Steam turbine regulating valve (GRE001VV) | 0 | normal |
| | | 1 | accidental all closed |
| | | 2 | accidental all opened |
| $B_{36}$ | Condenser vacuum pump (CVI101PO) | 0 | normal |
| | | 1 | accidental failure |
| $B_{37}$ | Status of three main steam isolation valves (VVP001/002/003VV) | 0 | normal |
| | | 1 | all closed |
| | | 0 | normal |
| $B_{38}$ | SG feedwater status | 1 | loss |
| | | 2 | moderate |